\documentclass[twoside,11pt]{article}
\pdfoutput=1
\usepackage{automl2020}
\usepackage{microtype}
\usepackage{graphicx}
\usepackage{subfigure}
\usepackage{booktabs} 
\usepackage{amsmath, amsfonts, bm}
\usepackage{cancel, bbm}
\usepackage{outlines}
\usepackage{xcolor}
\usepackage{algorithmic}
\usepackage[ruled]{algorithm2e}
\usepackage{ifthen}

\jmlrheading{Yufei Wang and Tianwei Ni}

\ShortHeadings{Meta-SAC}{Wang and Ni}
\firstpageno{1}

\begin{document}

\title{Meta-SAC: Auto-tune the Entropy Temperature of\\ Soft Actor-Critic via Metagradient}

\author{\name Yufei Wang\textsuperscript{1*}, \name Tianwei Ni\textsuperscript{2*}\\
\addr Carnegie Mellon University}

\maketitle

\newcommand{\yf}[1]{\textcolor{red}{Yufei: #1}}
\newcommand{\tw}[1]{\textcolor{blue}{TW: #1}}

\newcommand{\av}{\mathbf{a}}
\newcommand{\bv}{\mathbf{b}}
\newcommand{\cv}{\mathbf{c}}
\newcommand{\dv}{\mathbf{d}}
\newcommand{\ev}{\mathbf{e}}
\newcommand{\fv}{\mathbf{f}}
\newcommand{\gv}{\mathbf{g}}
\newcommand{\hv}{\mathbf{h}}
\newcommand{\iv}{\mathbf{i}}
\newcommand{\jv}{\mathbf{j}}
\newcommand{\kv}{\mathbf{k}}
\newcommand{\lv}{\mathbf{l}}
\newcommand{\mv}{\mathbf{m}}
\newcommand{\nv}{\mathbf{n}}
\newcommand{\ov}{\mathbf{o}}
\newcommand{\pv}{\mathbf{p}}
\newcommand{\qv}{\mathbf{q}}
\newcommand{\rv}{\mathbf{r}}
\newcommand{\sv}{\mathbf{s}}
\newcommand{\tv}{\mathbf{t}}
\newcommand{\uv}{\mathbf{u}}
\newcommand{\vv}{\mathbf{v}}
\newcommand{\wv}{\mathbf{w}}
\newcommand{\xv}{\mathbf{x}}
\newcommand{\yv}{\mathbf{y}}
\newcommand{\zv}{\mathbf{z}}
\newcommand{\entropy}{\mathcal{H}}
\newcommand{\ent}{\mathcal{H}}

\newcommand{\defeq}{\mathrel{\mathop:}=}
\newcommand{\Eb}[2]{\mathbb{E}_{#1}\left[#2\right]}

\newcommand{\policy}{\pi}
\newcommand{\policyold}{{\policy_\mathrm{old}}}
\newcommand{\policynew}{{\policy_\mathrm{new}}}
\newcommand{\policyme}{\pi_\mathrm{me}}
\newcommand{\policyopt}{{\policy^*}}
\newcommand{\paramsz}{{\theta_0}}
\newcommand{\params}{\theta}
\newcommand{\paramsi}{{\theta_i}}
\newcommand{\paramsip}{{\theta_{i+1}}}
\newcommand{\softpolicy}{{\policy_\mathrm{soft}}}
\newcommand{\hardpolicy}{{\policy^{\dagger}}}
\newcommand{\pparams}{{\phi}}   
\newcommand{\mparams}{{\psi}} 
\newcommand{\qparams}{{\omega}}   
\newcommand{\vparams}{{\psi}}   
\newcommand{\vtargetparams}{{\bar\psi}}   
\newcommand{\qtargetparams}{{\bar\theta}}   
\newcommand{\rb}{\mathcal D}
\newcommand{\rbkl}{\mathcal{D}_{KL}}
\newcommand{\env}{p_e}

\renewcommand{\cite}{\citep}

\newcommand\blfootnote[1]{%
  \begingroup
  \renewcommand\thefootnote{}\footnote{#1}%
  \addtocounter{footnote}{-1}%
  \endgroup
}

\blfootnote{\textsuperscript{1} yufeiw2@andrew.cmu.edu, \textsuperscript{2} tianwein@cs.cmu.edu, \textsuperscript{*} indicates equal contribution.}

\begin{abstract}
Exploration-exploitation dilemma has long been a crucial issue in reinforcement learning. 
In this paper, we propose a new approach to automatically balance between these two.
Our method is built upon the Soft Actor-Critic (SAC)~\cite{haarnoja2018soft} algorithm, which uses an ``entropy temperature" that balances the original task reward and the policy entropy, and hence controls the trade-off between exploitation and exploration. It is empirically shown that SAC is very sensitive to this \textit{hyperparameter}, and the follow-up work (SAC-v2)~\cite{haarnoja2018soft2}, which uses constrained optimization for automatic adjustment, has some limitations. 
The core of our method, namely Meta-SAC, is to use metagradient along with a novel meta objective to automatically tune the entropy temperature in SAC.
We show that Meta-SAC achieves promising performances on several of the Mujoco benchmarking tasks, and outperforms SAC-v2 over 10\% in one of the most challenging tasks, \texttt{humanoid-v2}.


\end{abstract}
\section{Introduction}

Reinforcement learning algorithms need to find a good balance between exploration and exploitation. On the one hand, the agent needs to explore the environment to gather useful information. On the other hand, it needs to exploit the knowledge it already obtains to improve its policy. 
There have been numerous works that aimed to tackle the exploitation/exploration trade-off issue. One category of methods use intrinsic rewards / count-based bonuses to enhance the exploration~\cite{ostrovski2017count, pathak2017curiosity}. Another family of methods use an entropy term to guide the agent to explore and exploit in a balanced way~\cite{mnih2016asynchronous, haarnoja2018soft}. Among those methods, Soft Actor-Critic (SAC)~\cite{haarnoja2018soft} achieves the state-of-the-art performance on Mujoco \cite{todorov2012mujoco}, a set of RL benchmarking continuous control tasks, and also performs well on lots of other robotics tasks.

SAC augments the traditional RL objective \cite{sutton2018reinforcement} with a policy entropy term: 
\begin{equation}
\label{eq:sac}
J(\pi) \defeq \sum_{t=0}^T\gamma^t \Eb{\sv_t, \av_t\sim \rho_{\pi}}{ r(\sv_t, \av_t) + \alpha \entropy(\pi(\cdot | \sv_t))}
\end{equation} 

This is also called the maximum entropy RL objective \cite{ziebart2008maximum, levine2018reinforcement}. It also defines a \textit{soft} version of value functions and bellman operators, where a new policy update rule is derived. 
The entropy temperature $\alpha$ plays a key role in the maximum entropy RL objective as it is the \textit{hyperparameter} that balances the expected original task reward and the expected policy entropy, thus balancing the exploitation versus exploration. 

SAC~\cite{haarnoja2018soft} (referred to as SAC-v1 throughout) is known to be particularly sensitive to the entropy temperature. For large temperature, the policy is encouraged to become nearly uniform and thus fails to exploit the reward signal, which substantially degrading the performance; for small temperature, though the policy learns quickly at first, it then becomes nearly deterministic and gets stuck at poor local minimal due to lack of exploration.
However, it is non-trivial to choose a proper value of the entropy temperature. The optimal value not only changes across different tasks, but also varies in the learning process as the policy improves.
In SAC-v1, this problem is solved by treating $\alpha$ as a hyperparameter and determining its value by grid search. This brings significant computational costs and manual efforts, and needs to be done for each new task.

Recently, gradient-based methods have been developed for hyperparameter optimization for deep neural networks~\cite{automl, feurer_hyperparameter_2019}. 
In a follow-up work of SAC-v1~\cite{haarnoja2018soft2} (referred to as SAC-v2 throughout), the authors derive a formula to adjust the value of $\alpha$ automatically in the learning process by adding an entropy constraint upon the original RL objective, turning it into a constrained optimization problem, and using dual gradient descent to solve it. 
However, the derived update rule has several drawbacks, among which the most severe one is that it introduced another hyperparameter ``target entropy'', which by itself needs to be tuned for each task. The authors give a heuristic formula for choosing this new hyperparameter, which performs empirically well on Mujoco tasks, however, it remains unknown whether it is the optimal choice for every task. 

In this paper, we propose to leverage the metagradient method~\cite{xu2018meta, zheng2018learning} to automatically tune the value of $\alpha$ during the learning process. In contrast to constrained optimization in SAC-v2, metagradient method introduces no more adaptive hyperparameter.\footnote{An adaptive hyperparameter is defined as the one that needs to be tuned for every task.} 
It is also a different and principled way of adjusting $\alpha$, where the value of $\alpha$ (thus the balance between exploration and exploitation) is optimized towards minimizing a \textit{meta loss}. 
In the early metagradient works, the meta loss is mostly defined as the policy gradient loss after the current policy update. In this paper, we propose to use a different meta loss to be consistent with the evaluation metric (i.e., the classic RL objective). The new meta loss lies between the DDPG~\cite{lillicrap2015continuous} and SAC objective, and we empirically verify its effectiveness.

In summary, the main contributions of this paper are as follows: 
We propose and derive a new method to automatically tune the entropy temperature $\alpha$ in SAC, which is based on metagradient and a novel meta loss, and does not introduce any adaptive hyperparameters. We show that the proposed method achieves state-of-the-art performance on several of the Mujoco locomotion tasks \cite{todorov2012mujoco}.

\section{Preliminaries}
\subsection{Metagradient}
\label{sec:metag}
Metagradient \cite{xu2018meta, zheng2018learning} is a general method for adapting some of the hyperparameters in the learning algorithm online. We follow the notations in a recent paper ~\cite{zahavy2020self} to give a brief review on it. 
Let $\theta$ denotes all the learnable parameters of the algorithm, e.g., the weights for the policy and value network, and $\zeta$ denotes all the hyperparameters, e.g., the learning rate of the optimizer, the discount factor $\gamma$, and the entropy temperature $\alpha$ in SAC. Let $\eta$ be a subset of the hyperparameters $\zeta$ that we want to adapt during the learning process. We call $\eta$ the metaparameters.
The update of the learnable parameters $\theta$ at step $t$ is done by optimizing them w.r.t. a \textbf{learning loss} that depends both on $\theta$ and $\eta$:
\begin{equation}
    \theta_{t+1}(\eta_t) \gets \theta_t - \lambda_{\theta} \nabla_{\theta}L_{learn}(\theta_t, \eta_t)
    \label{eq:learning loss}
\end{equation}

The metagradient method adapts the value of $\eta$ by optimizing them w.r.t. a \textbf{meta loss}: 
\begin{equation}
    \eta_{t+1} \gets \eta_t - \lambda_{\eta}\nabla_{\eta}L_{meta}(\theta_{t+1}(\eta_t))
    \label{eq:meta loss}
\end{equation}
where $\lambda_\theta$ and $\lambda_\eta$ are the respective learning rates. Note that the learnable parameters $\theta$ and the metaparameter $\eta$ are updated in an alternative fashion iteratively. 

\subsection{SAC}

SAC \cite{haarnoja2018soft} augments the standard RL objective with expected entropy of the policy by $J(\pi) \defeq \sum_{t=0}^T\gamma^t \Eb{\sv_t, \av_t\sim \rho_{\pi}}{ r(\sv_t, \av_t) + \alpha \entropy(\pi(\cdot | \sv_t))}$. 

SAC is an off-policy algorithm, where it stores a collection of $\{(\sv_t, \av_t, \rv_t, \sv_{t+1}\}_{i=1}^N$ transition tuples in a replay buffer $\rb$. It uses a neural network with parameter $\pparams$ for the policy as $\pi_\pparams$, and uses another neural network with paramter $\qparams$ for the Q-value as $Q_\qparams$. For training $\pparams$ and $\qparams$, it randomly samples a batch of transition tuples from the replay buffer, and performs stochastic gradient descent on minimizing the following loss objectives for $\pparams$ and $\qparams$:
\begin{equation}
\begin{split}
       L_Q(\qparams) &\defeq  \Eb{\sv_t, \av_t \sim \rb}{\frac12(Q_\qparams(\sv_t, \av_t) - Q^{\text{tar}}(\sv_t, \av_t))^2} 
    \\ \mathrm{where}\ 
        Q^{\text{tar}}(\sv_t, \av_t) &\defeq r(\sv_t, \av_t) + 
        \gamma \mathbb{E}_{\substack{\sv_{t+1}\sim \env\\ \av_{t+1}\sim \pi_\pparams}} [\hat{Q}(\sv_{t+1}, \av_{t+1}) - \alpha\log(\pi_\pparams(\av_{t+1}| \sv_{t+1}))]
\end{split}
\label{eq:sac-q-obj}
\end{equation}
\begin{equation}
    L_\policy(\pparams) \defeq \Eb{\sv_t \sim \rb, \av_t \sim \pi_\pparams} {\alpha\log\pi_\pparams(\av_t|\sv_t) - Q_\qparams(\sv_t, \av_t)}
    \label{eq:sac-pi-obj}
\end{equation}
where $\hat{Q}$ is target Q function whose parameters are periodically copied from the learned $Q_\qparams$.


SAC-v2~\cite{haarnoja2018soft2} makes the first step towards automating the tuning of $\alpha$ online. It casts the policy entropy term into a constraint that requires the  policy to have a minimal expected entropy: 
\begin{equation}
\begin{split}
    &\max_{\pi_{0:T}}\quad \Eb{\sv_t, \av_t\sim \rho_\pi}{\sum_{t=0}^T r(\sv_t, \av_t)}  \quad \text{s.t.}\quad \Eb{\sv_t,\av_t\sim \rho_\pi}{-\log(\pi_t(\av_t|\sv_t))} \geq H ~~\forall t
\end{split}
\end{equation}
$\alpha$ then becomes the dual variable in the dual problem. The optimal policy to this problem is time-varying. The authors derived an approximated update formula w.r.t. $\alpha$ using dual gradient descent on $L(\alpha) \defeq \Eb{\sv_t\sim \rb, \av_t\sim\rho_\pi}{-\alpha \log \pi(\av_t|\sv_t) - \alpha H}$ and dropping the time dependencies.
This update formula is empirically proved to work well on Mujoco tasks.

However, there are some issues with it. 
First, it has several key assumptions that is generally not true in real applications. For example, the derivation and convergence of the rule requires convexity, which does not hold for neural networks. The time dependency for the optimal solution is dropped for approximation. 
Second, for updating $\alpha$, this formula introduces another hyperparameter $H$, the minimal expected entropy. The authors give an empirical formula for the value of $H=-\mathrm{dim}(\av)$, i.e. the negative dimension of the action space, but it remains unknown how this is derived and whether it would work for each task. 
Besides, it seems contradictory to introduce an extra adaptive hyperparameter in the way of trying to free the tuning of the original one.

In the next section, we will introduce our proposed method of using metagradient to automate the turning of $\alpha$, which has minimal assumptions and introduces no more adaptive hyperparameters.

\section{Meta-SAC}

We now demonstrate how the learning loss and meta loss is instantiated when applying metagradient to adapt the entropy temperature $\alpha$ in SAC. Using the notations in the subsection \ref{sec:metag}, the learnable parameters are $\theta = \{\pparams, \qparams\}$, and the metaparameter is $\eta = \{\alpha\}$. Therefore, the learning loss for $\theta$ are $L_Q(\qparams)$ and $L_\pi(\pparams)$, and Eq. \ref{eq:learning loss} becomes:
\begin{equation}
    \begin{split}
        \pparams_{t+1}(\alpha_t) &\gets \pparams_t - \lambda_\pparams \nabla_\pparams L_\pi(\pparams_t, \alpha_t) \\
        \qparams_{t+1}(\alpha_t) &\gets \qparams_t - \lambda_\qparams \nabla_\qparams L_Q(\qparams_t, \alpha_t) 
    \end{split}
    \label{eq:sac learning loss}
\end{equation}

The choice of meta loss is critical for updating the metaparameter $\eta$. Most of the previous works~\cite{zheng2018learning, xu2018meta, zahavy2020self} use policy gradient loss as the meta loss. However, in our initial experiments, we find that policy gradient loss performs poorly. 
Instead, we propose to use the following meta loss:
\begin{equation}
    \label{eq:meta_alpha}
    L_{meta}(\alpha_t) \defeq \Eb{\sv_0 \sim \rb_0}{ -Q_{\qparams_{t}}(\sv_0, \pi^{\text{det}}_{\phi_{t+1}(\alpha_t)}(\sv_0))}
\end{equation}
In this meta loss, $\pi^{\text{det}}_{\phi_{t+1}(\alpha_t)}$ denotes the deterministic version of the updated policy (e.g., when the policy is parameterized as a Gaussian, the deterministic version always chooses the mean of the Gaussian), and $\rb_0$ denotes a special replay buffer that stores the initial states of the environments. 

The main intuition is to make the meta loss \textit{consistent} with the \textbf{evaluation metric} $M(\pi)$, i.e. the standard RL objective that only considers the task reward, as this is the quantity of concern to us:

\begin{equation}
\label{eq:test}
    M(\pi) \defeq \Eb{\sv_t,\av_t \sim \rho_\pi}{\sum_{t=0}^T \gamma^t r(\sv_t,\av_t)} = \Eb{\sv_0\sim \env, \av_0\sim \pi}{Q^\pi(\sv_0, \av_0)}
\end{equation}
where $Q^\pi$ is the classic Q-value function for policy $\pi$ without considering the entropy term, and the initial state $\sv_0$ is sampled from the environment.
We now elaborate on how the evaluation metric leads to the design of our meta loss: 
\begin{enumerate}
\itemsep0em 

\item The evaluation metric $M(\pi)$ does not consider the policy entropy, thus we drop the entropy term in the SAC loss (Eq. \ref{eq:sac-pi-obj}). Moreover, normally during evaluation we cast the stochastic policy into a deterministic one, and this leads to the design of using $\pi^{\text{det}}_{\phi_{t+1}(\alpha_t)}$.
     
\item As shown in Eq. \ref{eq:test}, the evaluation metric is expectation over initial state distribution. Therefore, we collect and store some initial states into a special buffer $\rb_{0}$, and sample a batch of initial states $\sv_0$ from $\rb_{0}$ to train meta loss. This method performs much better than sampling from \textit{arbitrary} states, as shown by our ablation studies in appendix \ref{sec:s0}.
    
\item The evaluation metric is based on the classic Q-value that does not consider the entropy term. However, in our meta loss we still use the soft Q value as it helps exploration, and our experiments show that using the classic Q-value performs worse in appendix \ref{sec:q}. This is the key discrepancy between our meta loss and the DDPG loss~\cite{lillicrap2015continuous}. Our meta loss actually lies between the DDPG loss (we use soft-Q instead of classic Q) and the SAC loss (we drop the entropy term). 
     
\item We use the old soft Q function $Q_{\qparams_t}$ instead of the updated one $Q_{\qparams_{t+1}(\alpha_t)}$. The main reason is that taking derivative of $Q_{\qparams_{t+1}(\alpha_t)}(\sv, \pi^{\text{det}}_{\phi_{t+1}(\alpha_t)}(\sv))$ w.r.t. $\alpha$ is very numerically unstable in our early experiments.

\end{enumerate}
The full algorithm of Meta-SAC is provided in appendix \ref{sec:code}.
 


\section{Experiments}

We compare Meta-SAC with three state-of-the-art off-policy RL algorithms, including SAC-v1, SAC-v2, and TD3 \cite{fujimoto2018addressing} which made several improvements upon DDPG \cite{lillicrap2015continuous}. The main competitor is SAC-v2, as it is the only existing method that automatically tunes the value of $\alpha$. 

Our experiments aim to answer the following questions: (1) How does Meta-SAC perform across a range of different environments? (2) How does the entropy temperature $\alpha$ evolve during the learning process? (3) How effective is the proposed new meta objective?\footnote{Due to space limit, the third question is answered in appendix \ref{sec:ablation} as ablation studies.}

Our source code is available online\footnote{\url{https://github.com/twni2016/Meta-SAC}}.
All the hyperparameters are listed in appendix \ref{sec:hp}. 

\subsection{Mujoco Learning Curves}
To answer the first question, we select four representative environments from the Mujoco benchmarking continuous control tasks \cite{todorov2012mujoco}, namely \texttt{Ant-v2, Hopper-v2, Humanoid-v2, Walker2d-v2}. The environment details can be found in appendix \ref{sec:env}. 
Among them, \texttt{Humanoid-v2} is one of the most challenging tasks that can be solved by current RL algorithms \cite{salimans2017evolution}, with a very high-dimensional state space (376). 

\begin{figure}[h]
    \centering
    \includegraphics[width=0.24\textwidth]{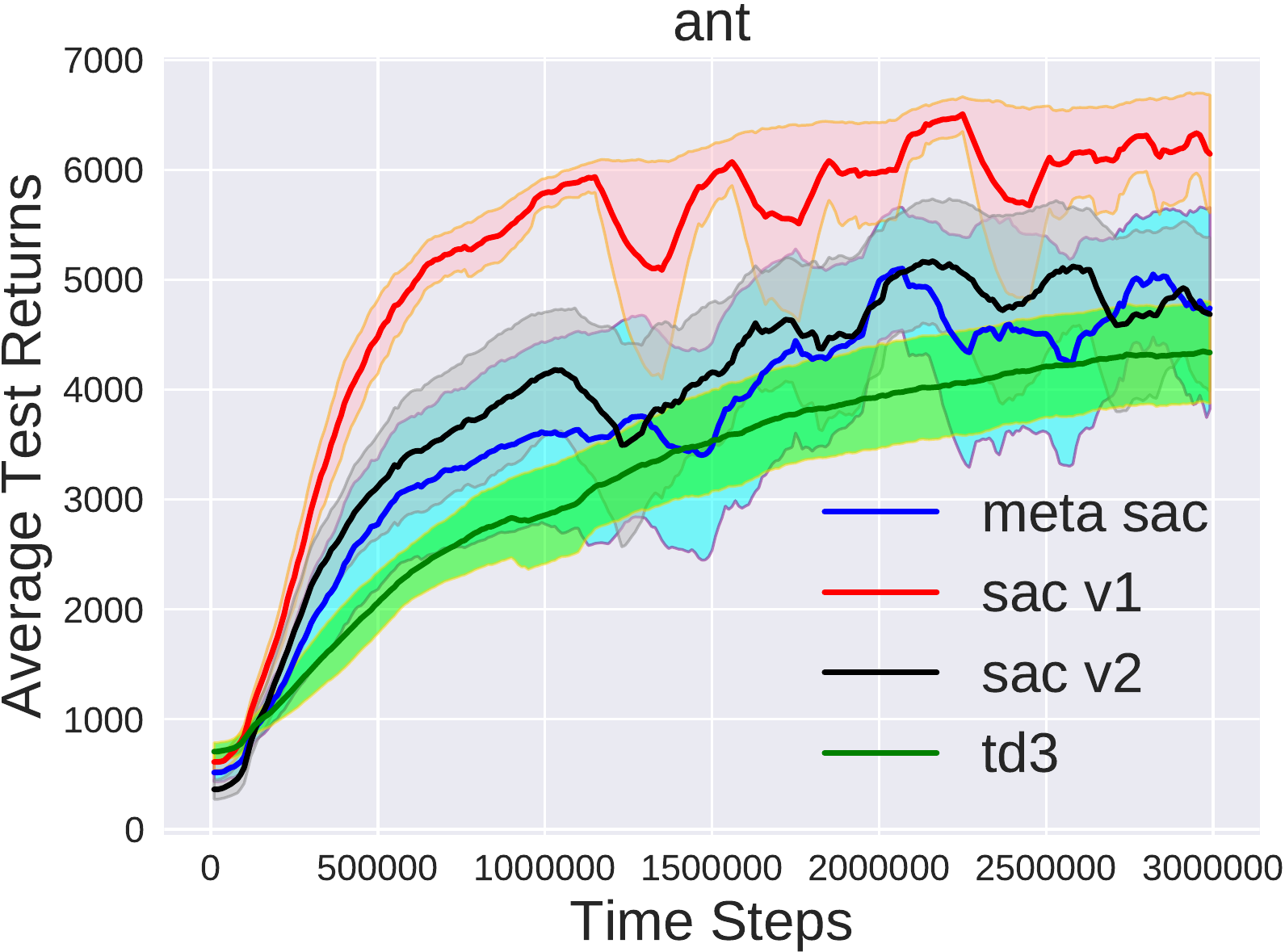}
    \includegraphics[width=0.24\textwidth]{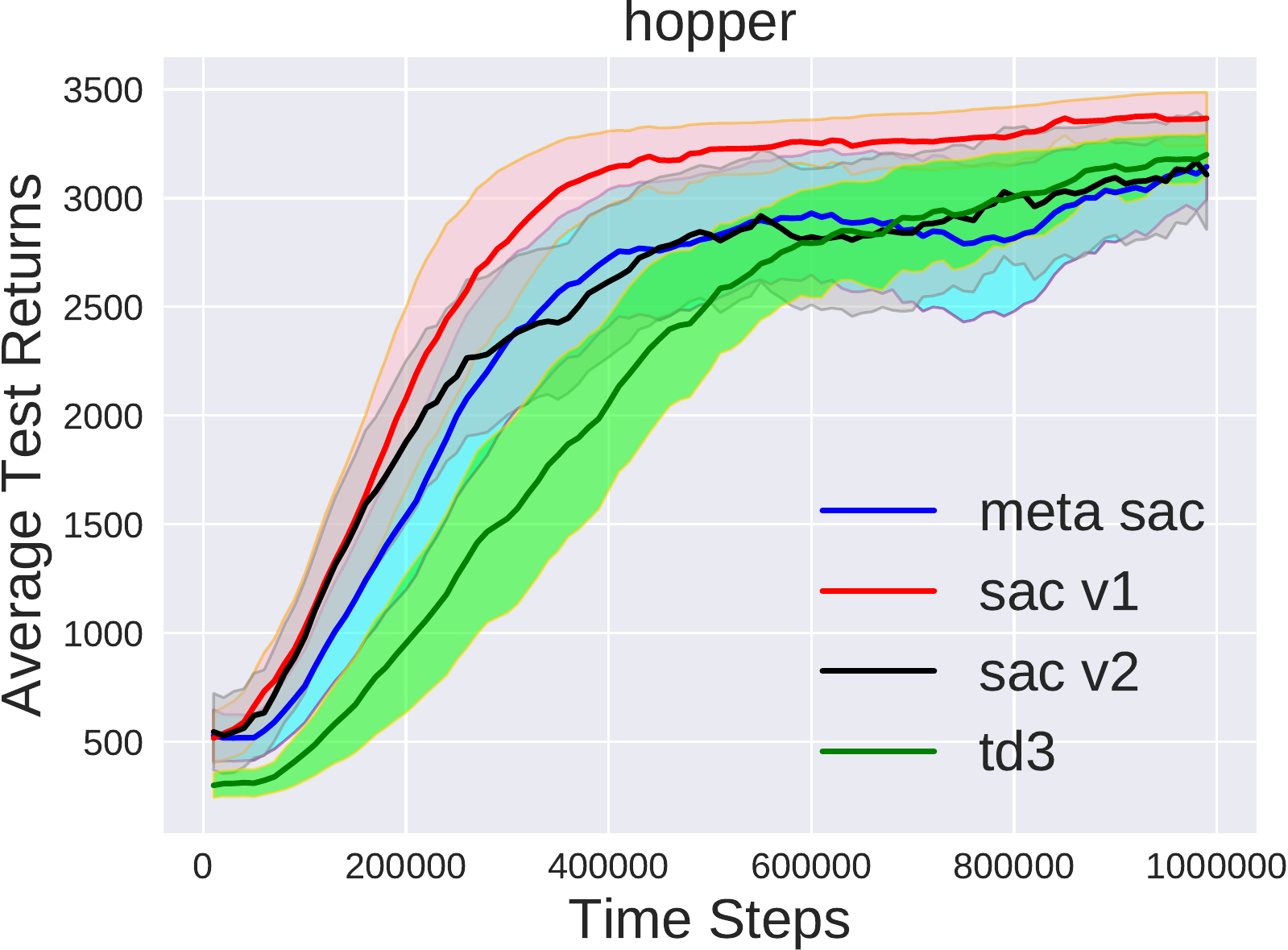}
    \includegraphics[width=0.24\textwidth]{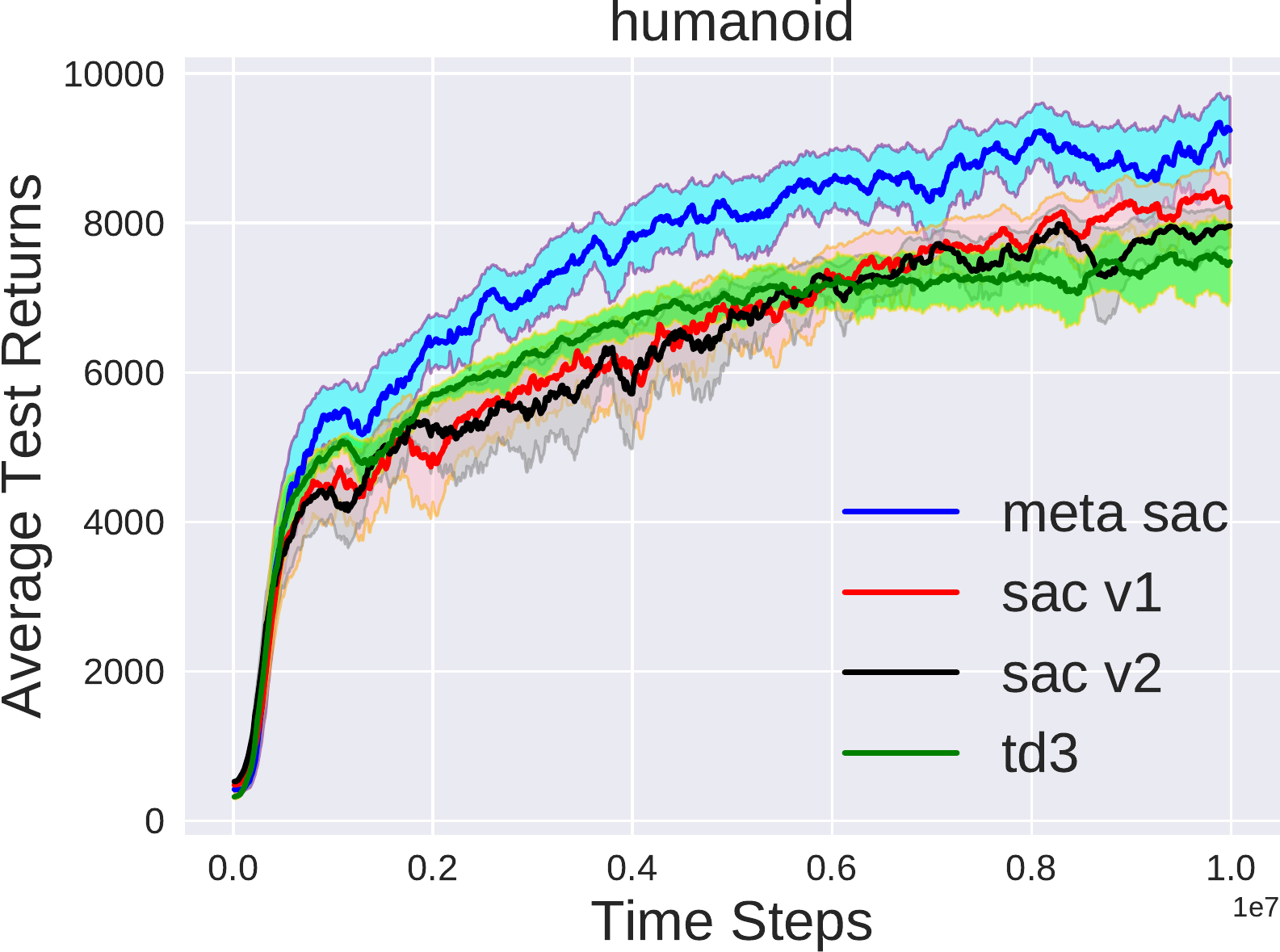}
    \includegraphics[width=0.24\textwidth]{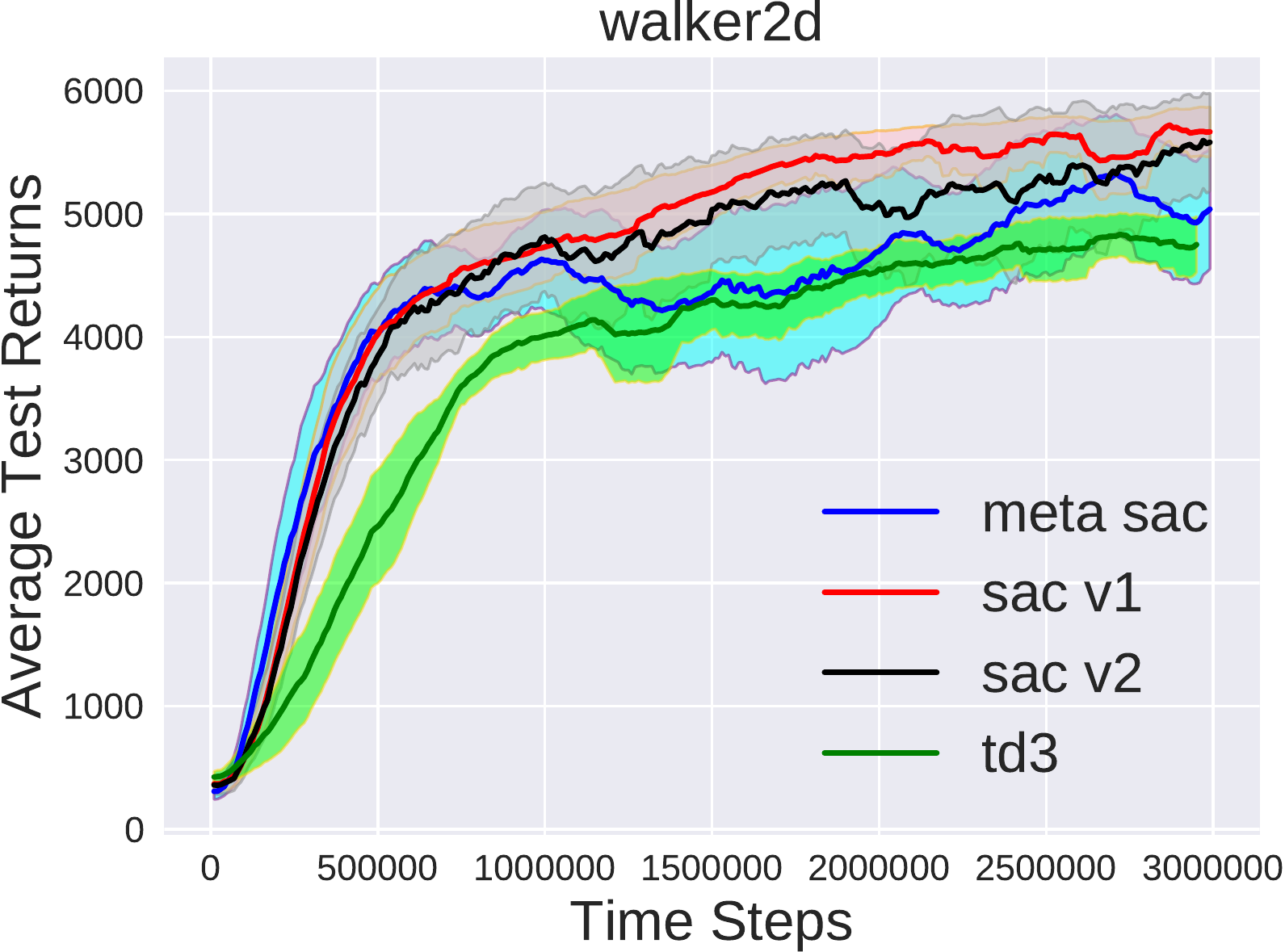}
    \includegraphics[width=0.24\textwidth]{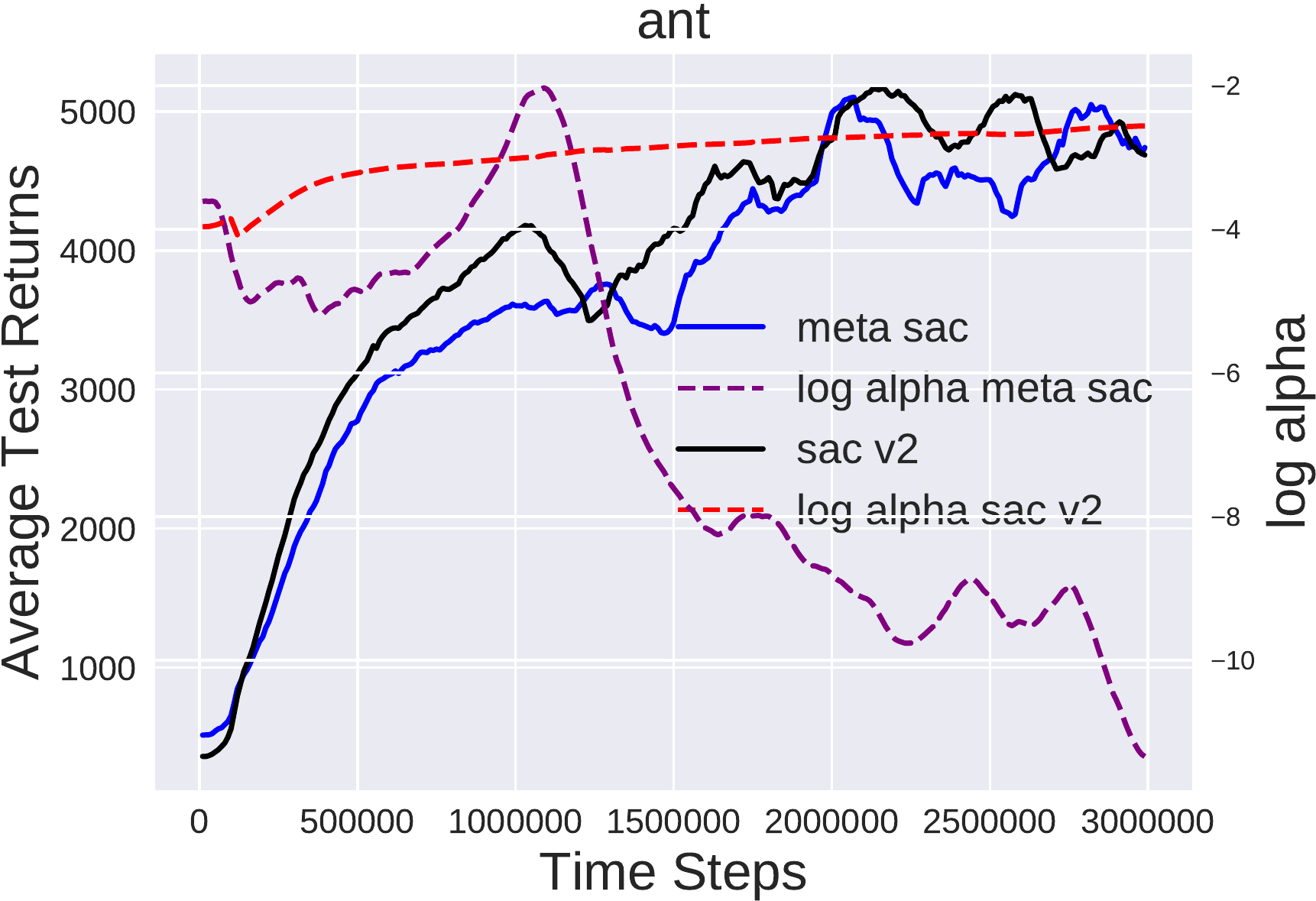}
    \includegraphics[width=0.24\textwidth]{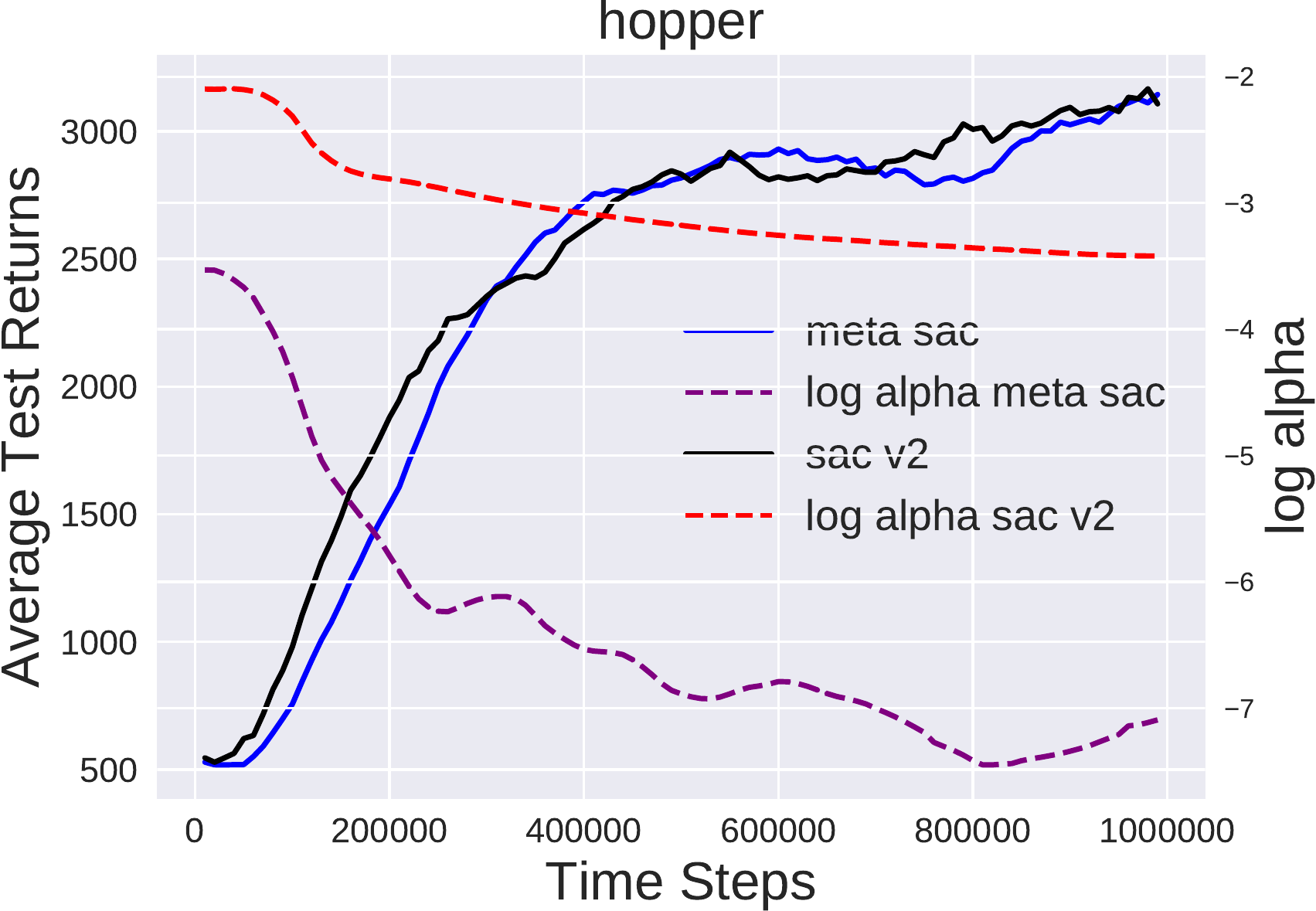}
    \includegraphics[width=0.24\textwidth]{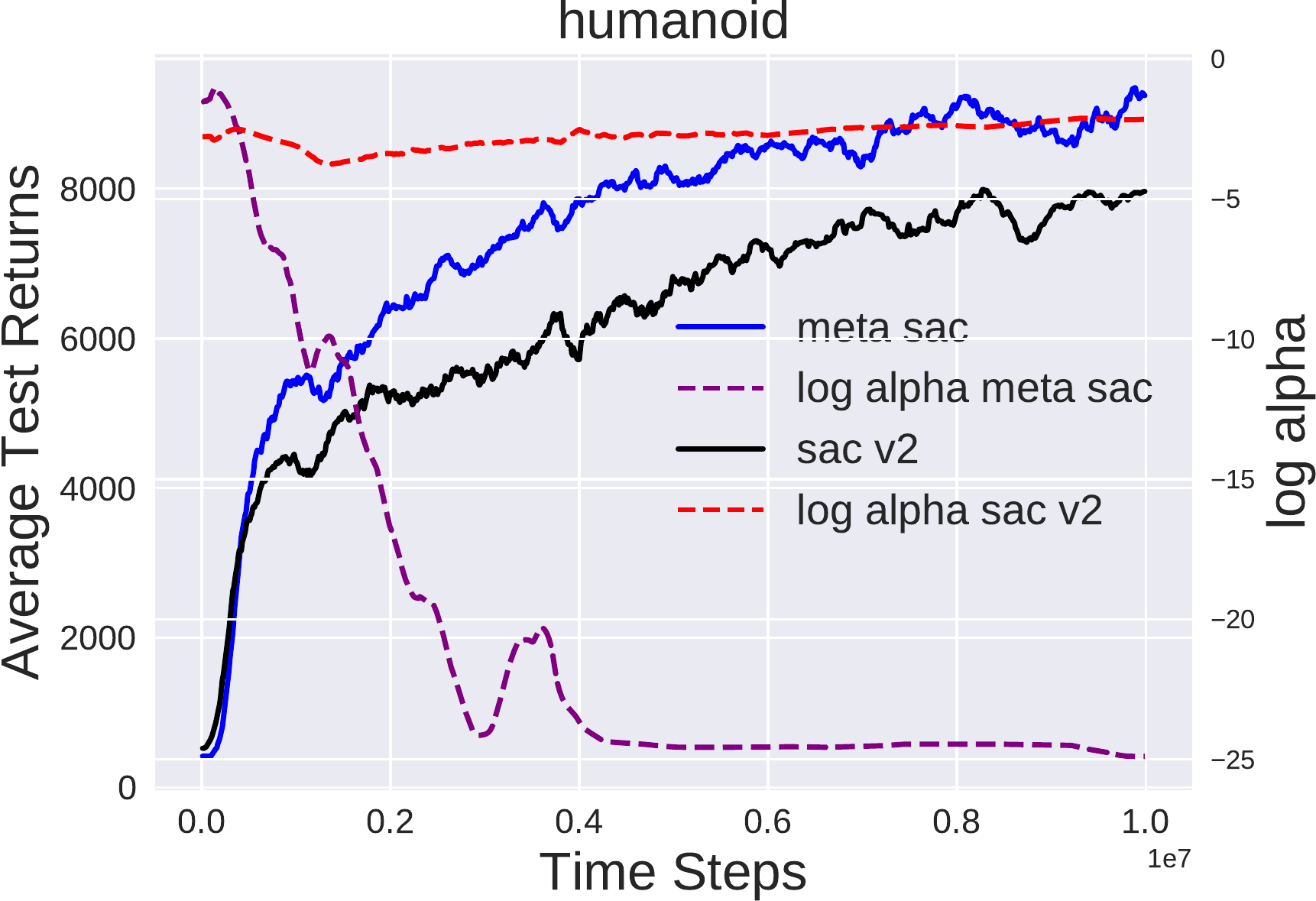}
    \includegraphics[width=0.24\textwidth]{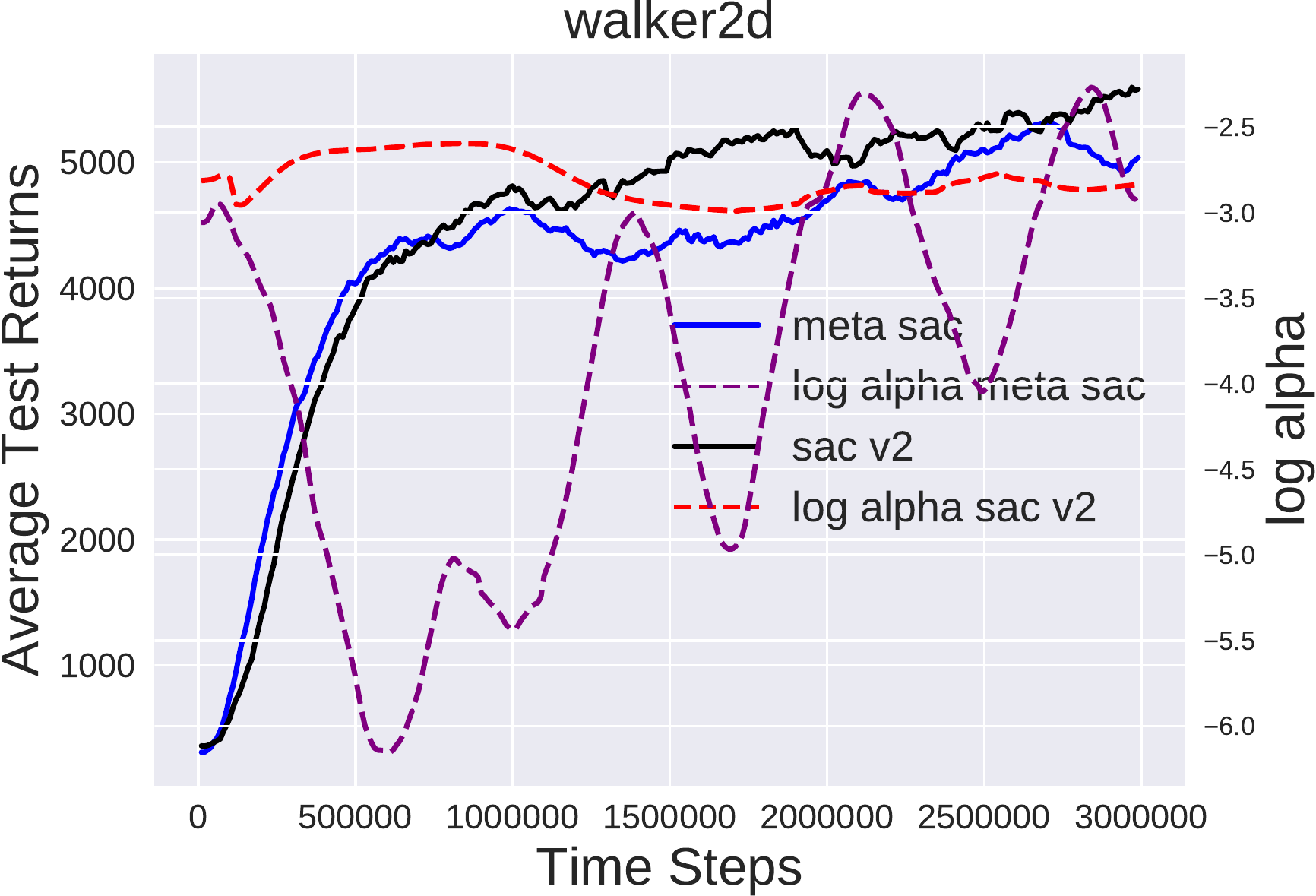}
    \caption{The first row shows the average test returns during the training process on the 4 Mujoco tasks of all compared algorithms. The curves show the mean and the shaded regions show half of the std over 5 random seeds. The curves are smoothed over the last 20 evaluations for better visualization. The second row shows the average test returns (solid) and the corresponding $\log \alpha$ curves (dashed) for the 4 tasks during the training process in SAC-v2 and Meta-SAC. }
    \label{fig:learn_curve}
\end{figure}

The first row of Figure \ref{fig:learn_curve} shows the average return of evaluation rollouts through training process for Meta-SAC (blue), SAC-v1 (red), SAC-v2 (black) and TD3 (green). We follow the evaluation standard of SAC-v2: we run five different instances of each algorithm with different random seeds, do 10 evaluation rollouts every 10000 training environment steps, and report the mean of the return of these 10 rollouts. As shown in the figure,  Meta-SAC achieves comparable performance with SAC-v2 on the easy tasks (\texttt{Ant-v2,Hopper-v2,Walker2d-2}), and performs slightly worse than SAC-v1 which uses grid search for the value of $\alpha$ for each task. In the most difficult task \texttt{Humanoid-v2}, Meta-SAC performs significantly better than all the other methods. It not only achieves a final return that is $10\%$ higher than the others, but also demonstrates faster convergence. These results show that Meta-SAC can be a strong alternative to SAC-v2, especially for 
complex tasks.

\subsection{How Entropy Temperature Changes during Learning?}
\label{sec:alpha}
The second row of Figure \ref{fig:learn_curve} shows how differently $\log \alpha$ changes during the training process between SAC-v2 and Meta-SAC. In SAC-v2, the change of $\log \alpha$ is \textit{mild}, which reaches a plateau after just a few learning steps. 
On the contrast, $\log \alpha$ changes dramatically in Meta-SAC along with a much larger scale. Generally in the later learning stages, $\alpha$ almost converges to zero, which means almost no entropy bonus is given and the SAC objective becomes almost \textit{equivalent} to the DDPG objective. 
This might explain why Meta-SAC performs much better in the \texttt{Humanoid-v2}, as large entropy could indeed help exploration in the early stage of the training, but in the later stages when the policy has been trained well, a smaller exploration bonus would help the policy exploit better what it already learned. We verify this by decaying the temperature in SAC-v1 shown in appendix \ref{ab:sac_v1}. This phenomenon also matches how DQN-style method adjusts the value of $\epsilon$ in $\epsilon$-greedy strategy, where it gradually decays to zero during the training.

\section{Conclusion}
In this paper, we present an auto-tuning method on the entropy temperature of the SAC algorithm using metagradient, with a novel meta loss aimed to be consistent with the evaluation metric. We verify its effectiveness in Mujoco benchmarking tasks, where it achieves state-of-the-art performance on one of the most difficult tasks, \texttt{Humanoid-v2}.


\acks{
The authors thank Shicong Cen and Jiaqiang Ruan for the discussion in the early experiments. We also thank CMU 10-715 class lecturer Nihar B. Shah and the classmates for their suggestions in the early version. We thank the anonymous reviewers whose comments helped improve and clarify this paper.
}

\bibliography{ref}


\newpage

\appendix
\clearpage

\section{Meta SAC Algorithm}
\label{sec:code}
\begin{algorithm}
\caption{Meta-SAC}
\label{alg:meta-sac}
\begin{algorithmic}
\STATE \mbox{Initialize Q network parameters $\qparams_0$, policy network parameters $\pparams_0$, and alpha $\alpha_0$.} \\
\STATE \mbox{Empty replay buffer $\rb$, learning rates $\lambda_\qparams, \lambda_\pparams,\lambda_\alpha$, batch size $B$, start step $T_S$ }\\
\STATE \mbox{Collect a buffer $\mathcal{D}_0$ of initial states $\sv_0$ from environment resets}

\For{each training timestep $t$}{
        $\av_t \sim \policy_\pparams(\av_t |\sv_t )$ \\
        $\sv_{t+1} \sim p(\sv_{t+1}| \sv_t, \av_t)$ \\
        $\mathcal{D} \leftarrow \mathcal{D} \cup \left\{(\sv_t, \av_t, r(\sv_t, \av_t), \sv_{t+1}) \right\}$\\
        \If{$t>T_S$}{
        \STATE \mbox{Sample a batch of transitions $\mathcal{B} = \left\{(\sv, \av, r(\sv, \av), \sv') \right\}_{i=1}^{B}$} from $\mathcal{D}$. \\
        \item $\qparams_{t+1} \leftarrow \qparams_t - \lambda_\qparams \hat \nabla_\qparams L_Q(\qparams, \alpha_t)$ using $\mathcal{B}$ and Eq. \ref{eq:sac-q-obj}\\
        \STATE $\pparams_{t+1} \leftarrow \pparams_t - \lambda_\pparams \hat \nabla_{\pparams} L_\pi(\pparams_t, \alpha_t)$ using $\mathcal{B}$ and Eq. \ref{eq:sac-pi-obj} \\
        \STATE $\alpha_{t+1} \leftarrow \alpha_t - \lambda_\alpha L_{meta}(\alpha_t)$ using $\mathcal{D}_0$ and Eq. \ref{eq:meta_alpha} \\
        }
}
\end{algorithmic}
\end{algorithm}

In practice, we do resampling on a new minibatch $\mathcal{B}'$ from replay buffer $\rb$ for updating Q value $Q_\omega$ and policy $\pi_\phi$ to avoid overfitting on the minibatch $\mathcal{B}$ used for metaparameter $\alpha$ updates. This trick is also applied in \cite{finn2017model, zheng2018learning} for online cross-validation. The ablation study on resampling is in appendix \ref{sec:resample}.

\section{Environment Details}
\label{sec:env}
We carry out experiments on the locomotion tasks of simulated robots in Mujoco environments \cite{todorov2012mujoco} wrapped in OpenAI Gym \cite{brockman2016openai}. The states are the robots' generalized positions and velocities, and the actions are joint torques. 
The reward is defined to be proportional to the speed regularized with the acceleration in order to make locomotion more smooth. Early termination is added when the robot falls over, determined by thresholds on the height and torso angles. 
The locomotion tasks are challenging due to the high-dimensional state and action space, non-smooth dynamics, and underactuated systems \cite{schulman2015trust, lillicrap2015continuous}. 

\begin{figure}[h]
    \centering
    \includegraphics[width=0.3\textwidth]{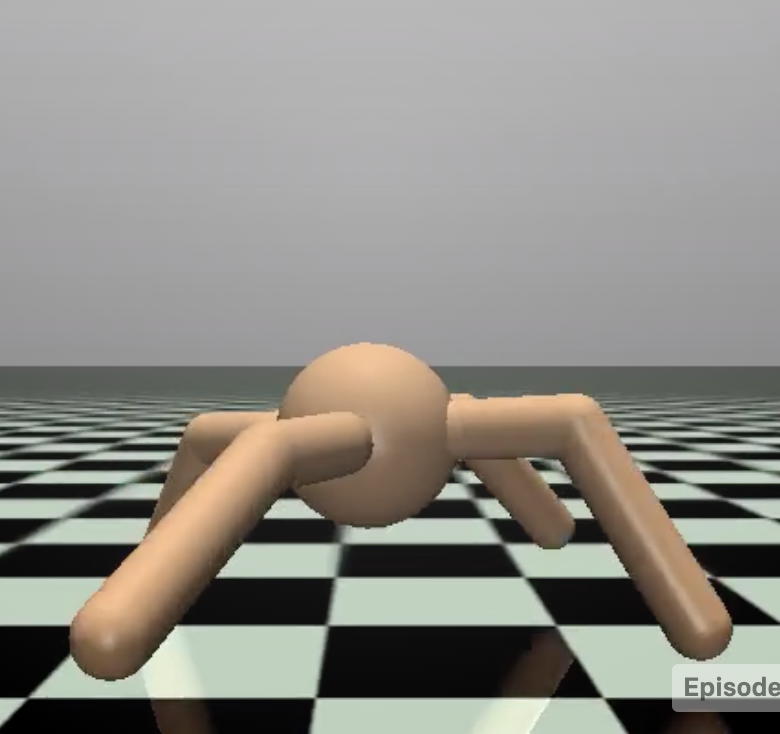}
    \includegraphics[width=0.3\textwidth]{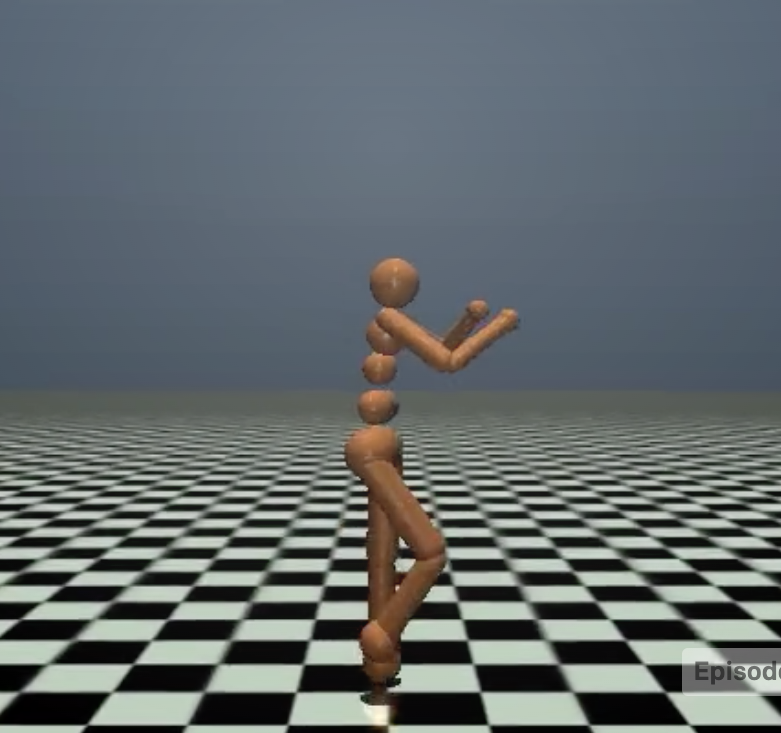}
    \caption{Rendering of simulated environments in \texttt{Ant-v2} (left) and \texttt{Humanoid-v2} (right).}
    \label{fig:gym}
\end{figure}

Figure \ref{fig:gym} shows the rendering of two of the locomotion tasks, \texttt{Ant-v2} and \texttt{Humanoid-v2} at one frame (figure sources\footnote{\url{https://gym.openai.com/envs/Ant-v2/} and \url{https://gym.openai.com/envs/Humanoid-v2/}}).

Table \ref{tab:gym} shows the dimensionality on state ($\textrm{dim}(\sv)$) and action ($\textrm{dim}(\av)$), along with a brief description of the tasks \cite{brockman2016openai} . The state space is unbounded, and the action space is in general bounded within $(-1, 1)$ in each dimension, except for \texttt{Humanoid-v2} whose action is bounded within $(-0.4, 0.4)$.

\begin{table}[h]
    \centering
    \begin{tabular}{c|c|c|c}
       \textbf{Task name}  & $\textbf{dim}(\sv)$ & $\textbf{dim}(\av)$ & \textbf{Description} \\ \hline
     \texttt{Ant-v2}       & 111 & 8 & Make a 3D four-legged robot walk forward as fast as possible. \\ 
     \texttt{Hopper-v2}    & 11  & 3 & Make a 2D one-legged robot hop forward as fast as possible. \\ 
     \texttt{Humanoid-v2}  & 376 & 17& Make a 3D bipedal robot walk forward as fast as possible. \\ 
     \texttt{Walker2d-v2}  & 17  & 6 & Make a 2D bipedal robot walk forward as fast as possible. \\ 
    \end{tabular}
    \caption{Brief information of the locomotion tasks.}
    \label{tab:gym}
\end{table}

\section{Hyperparameters}
\label{sec:hp}

\subsection{SAC Hyperparameters}

For SAC-v1 and SAC-v2, we directly use their original hyperparameters (refer to Section D in the appendix of SAC-v2 \cite{haarnoja2018soft2}). Specifically, we parameterize the policy network $\theta$ and the Q network $\phi$ both with a MLP of two hidden layers, with $256$ neurons and the ReLU \cite{nair2010rectified} activation for each layer. The action output of the policy network is parameterized by a squashed Gaussian \cite{haarnoja2018soft} with the same output range as the action space, where the mean and the diagonal covariance matrix are learnable parameters. 

We use a batch size of $256$ and the Adam \cite{kingma2014adam} optimizer with a learning rate of $3\cdot 10^{-4}$. The discount factor $\gamma$ is $0.99$, the target smoothing coefficient $\tau$ is $0.05$, and the replay buffer size is $10^6$. The start step $T_S$ is $10000$ for each task.

Table \ref{tab:sac} shows the hyperparameters that are different for each task in the baseline SAC algorithms.

\begin{table}[h]
    \centering
    \begin{tabular}{c|c|c|c|c}
    & \texttt{Ant-v2} & \texttt{Hopper-v2} &
    \texttt{Humanoid-v2} & \texttt{Walker2d-v2}   \\\hline
    \textbf{temperature} $\bm{\alpha}$ for SAC-v1 & 0.2 & 0.2 & 0.05 & 0.2 \\
    \textbf{entropy target} $\bm{H}$ for SAC-v2 & -8 & -3 & -17 & -6 \\
    training timesteps (in \textit{Million}) & 3 & 1 & 10 & 3 
    \end{tabular}
    \caption{Hyperparameters that are different for each task in SAC.}
    \label{tab:sac}
\end{table}

\subsection{Meta-SAC Hyperparameters}
The hyperparameters that are different from SAC to Meta-SAC are listed as follows:
\begin{itemize}
\itemsep0em 
    \item We change the policy optimizer to RMSProp \cite{hinton2012neural} with $\epsilon=10^{-12}$ for better performance, and keep the same learning rate in Meta-SAC. Note that SAC performs similarly under both Adam and RMSProp. The main reason that we use RMSProp is that our implementation requires backpropagating through the update process of the optimizer, and the update rule of RMSProp is more numerically stable during backpropagation compared with Adam.
    \item The metaparameter, the entropy temperature $\alpha$ is parameterized in the form of $\log \alpha$, and we clip its value to be below zero. This ensures $0 < \alpha \le 1$.
    \item The learning rate for $\log \alpha$ is $3\cdot 10^{-4}$, which is kept to be the same as the learning rate of the policy or the Q function. Its gradient norm is clipped below $0.05$ to stabilize training.
    \item The size of replay buffer $\rb_0$ is same as batch size $256$.
\end{itemize}

It should be emphasized that though we introduce several extra hyperparameters, we keep all of them the same value across different tasks, whereas SAC-v1 and SAC-v2 have to choose different values of the entropy temperature $\alpha$ or the entropy target $H$ in each task.

\section{Ablation Studies on Meta Objective}
\label{sec:ablation}
In this section, we aim to answer the third question that we proposed in the main paper:
How effective is the proposed new meta objective? We do the ablative analysis under the same experiment setting as Meta-SAC except for the meta objective.

\subsection{Ablation Study on Using Initial States}
\label{sec:s0}
First we test the effectiveness of using the environment initial states instead of arbitrary states in the meta loss. The result is shown in the first row of Figure \ref{fig:ablation}. As can be seen, using the environment initial states significantly boost the performance of Meta-SAC, especially in \texttt{Ant-v2} and \texttt{Humanoid-v2}. 

\subsection{Ablation Study on Resampling}
\label{sec:resample}
Then we test how the Meta-SAC performs if we change the soft-Q value used in Eq. \ref{eq:meta_alpha} to the classic Q-value. The result is shown in Figure \ref{fig:ablation} second row. As can be seen, when using the classic Q-function in the meta-loss, the performances degrades obviously in \texttt{Ant-v2} and \texttt{Walker2d-v2}. We suspect the reason is that such a meta loss discourages exploration and thus hurts the performance. 

\subsection{Ablation Study on Using Soft-Q Value}
\label{sec:q}
\begin{figure}[h]
    \centering
    \begin{tabular}{ccc}
    \includegraphics[width=0.32\textwidth]{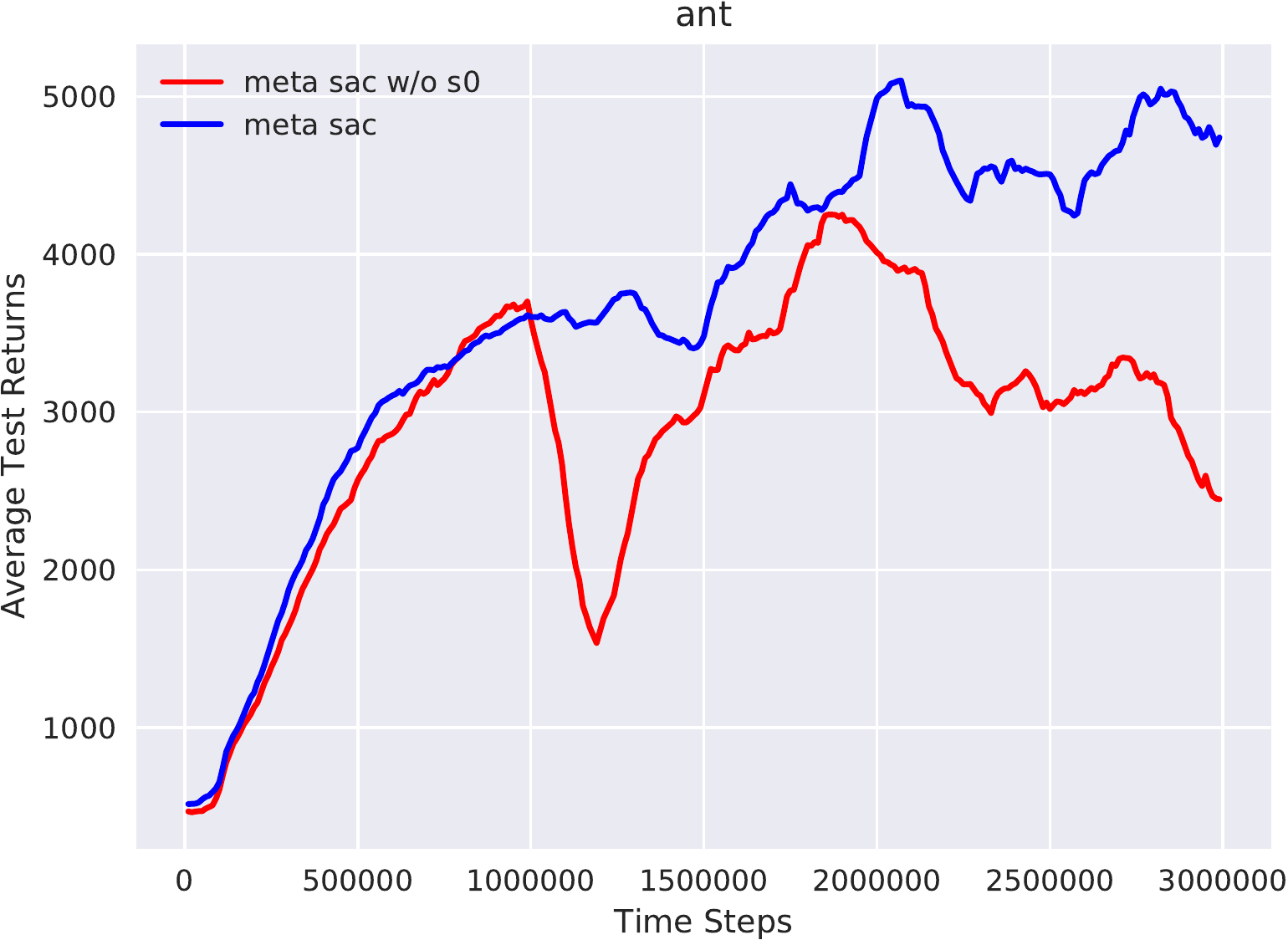} &
    \includegraphics[width=0.32\textwidth]{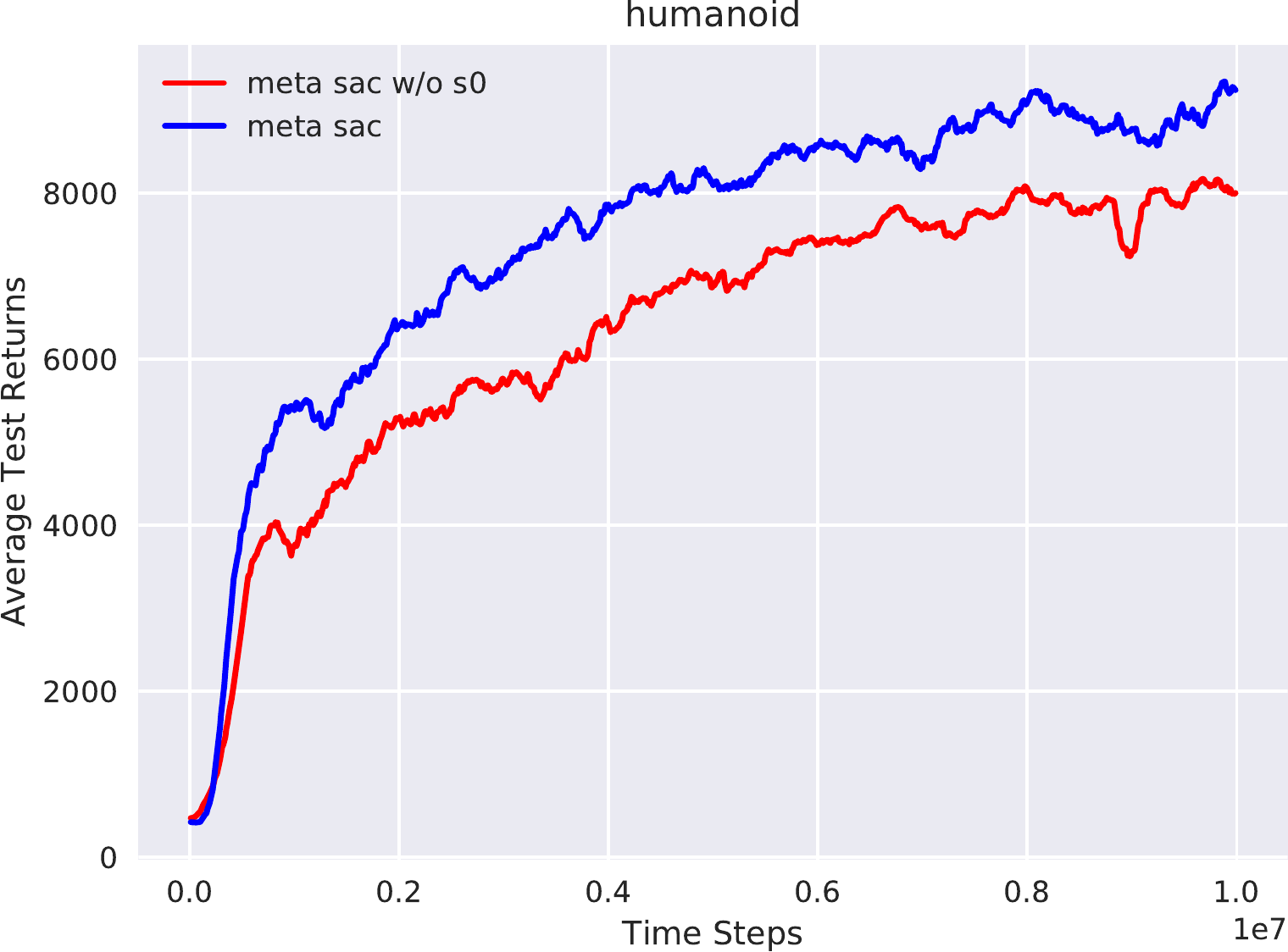} &
    \includegraphics[width=0.32\textwidth]{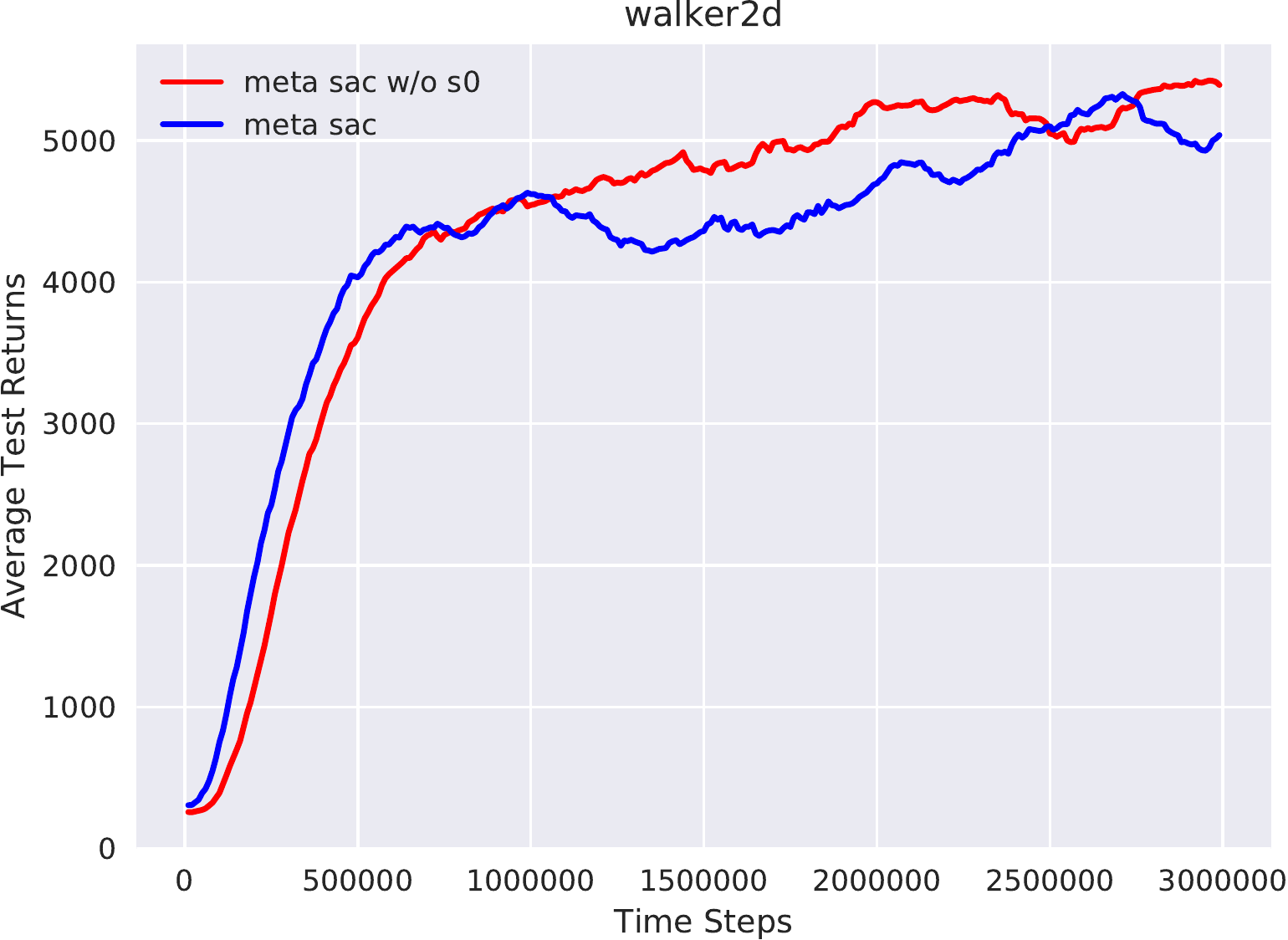} \\
    \multicolumn{3}{c}{(a) Ablation study on using arbitrary states (red curve) in the meta loss.}\\
    \includegraphics[width=0.32\textwidth]{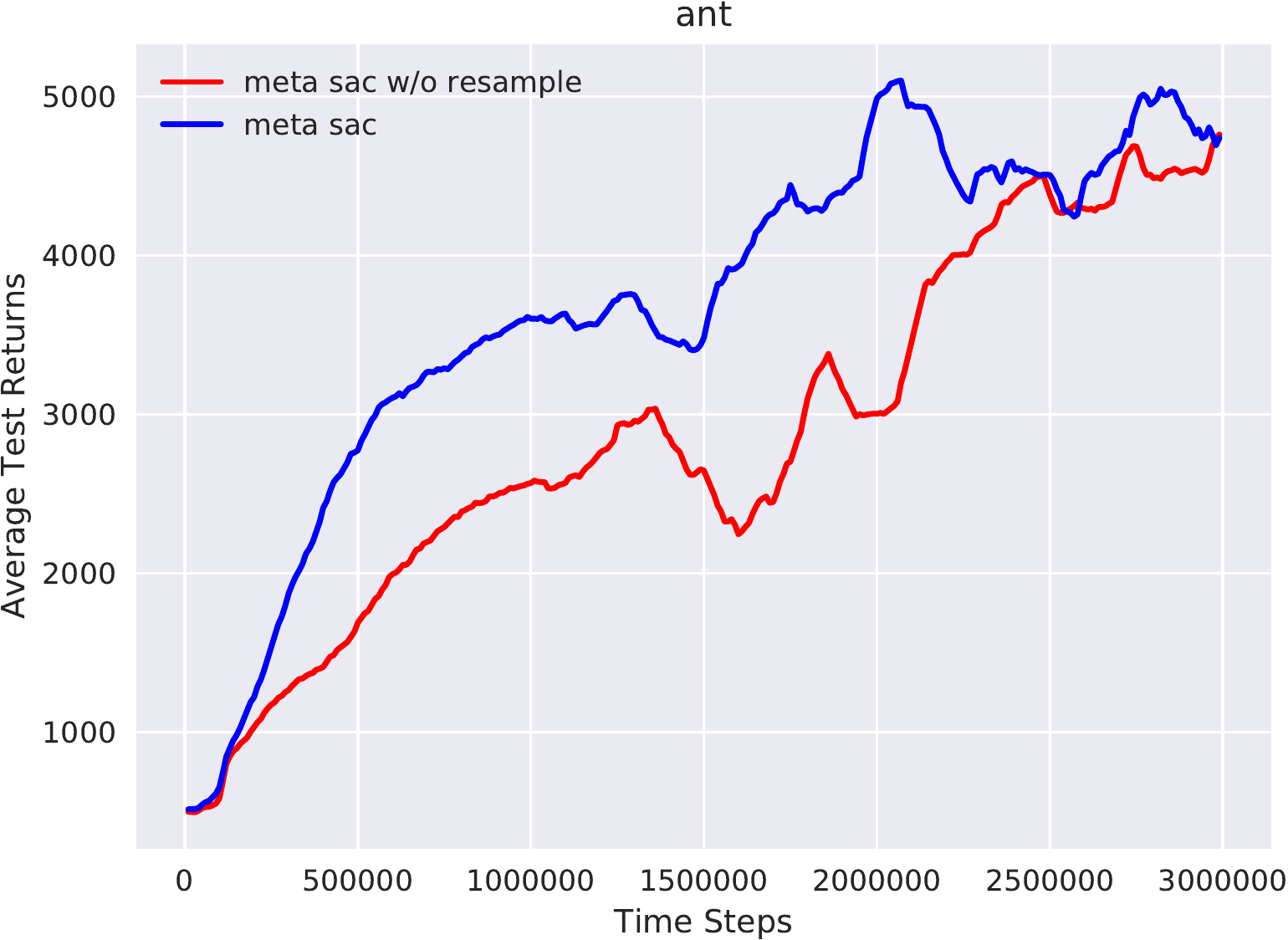} &
    \includegraphics[width=0.32\textwidth]{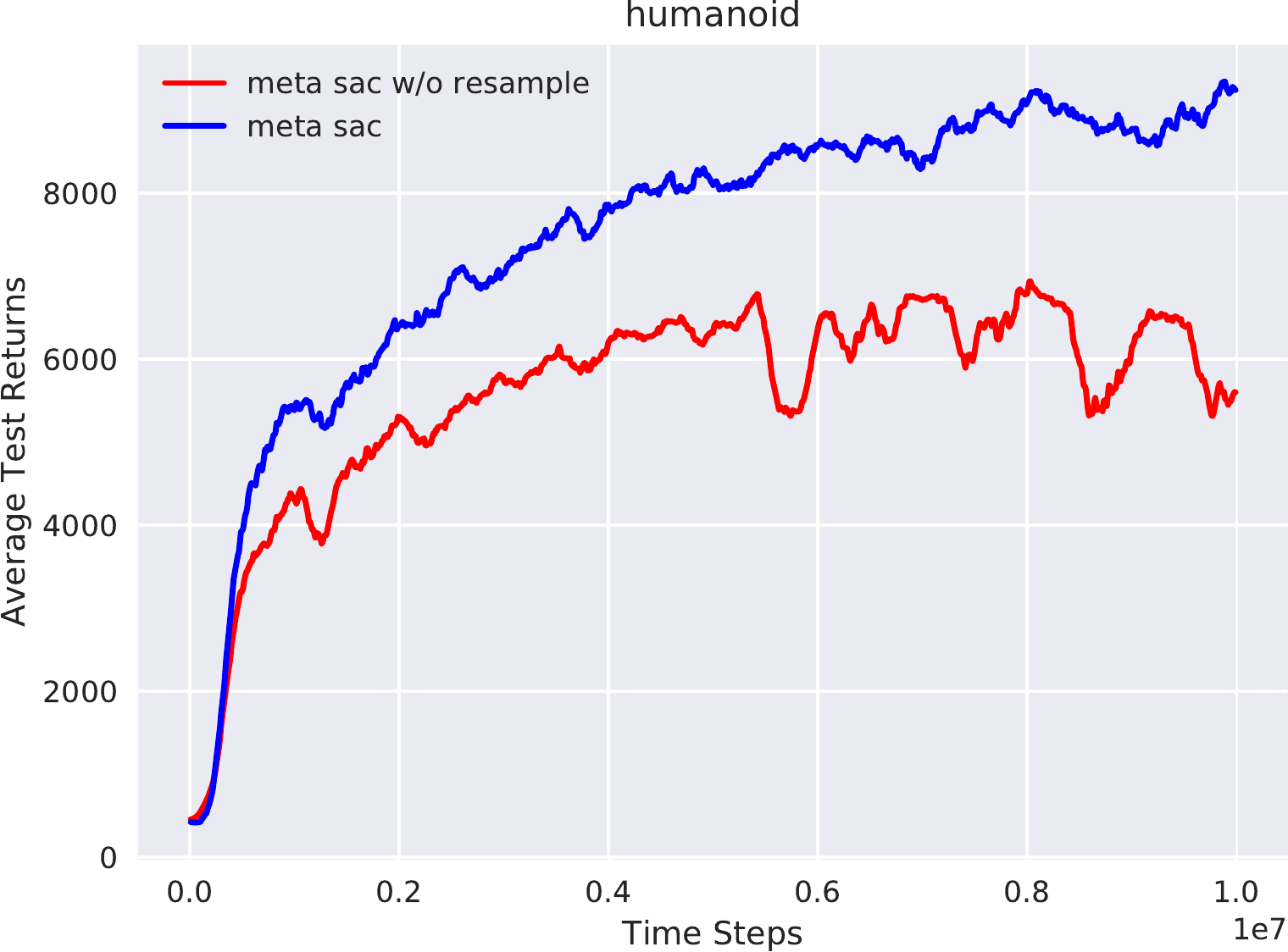} &
    \includegraphics[width=0.32\textwidth]{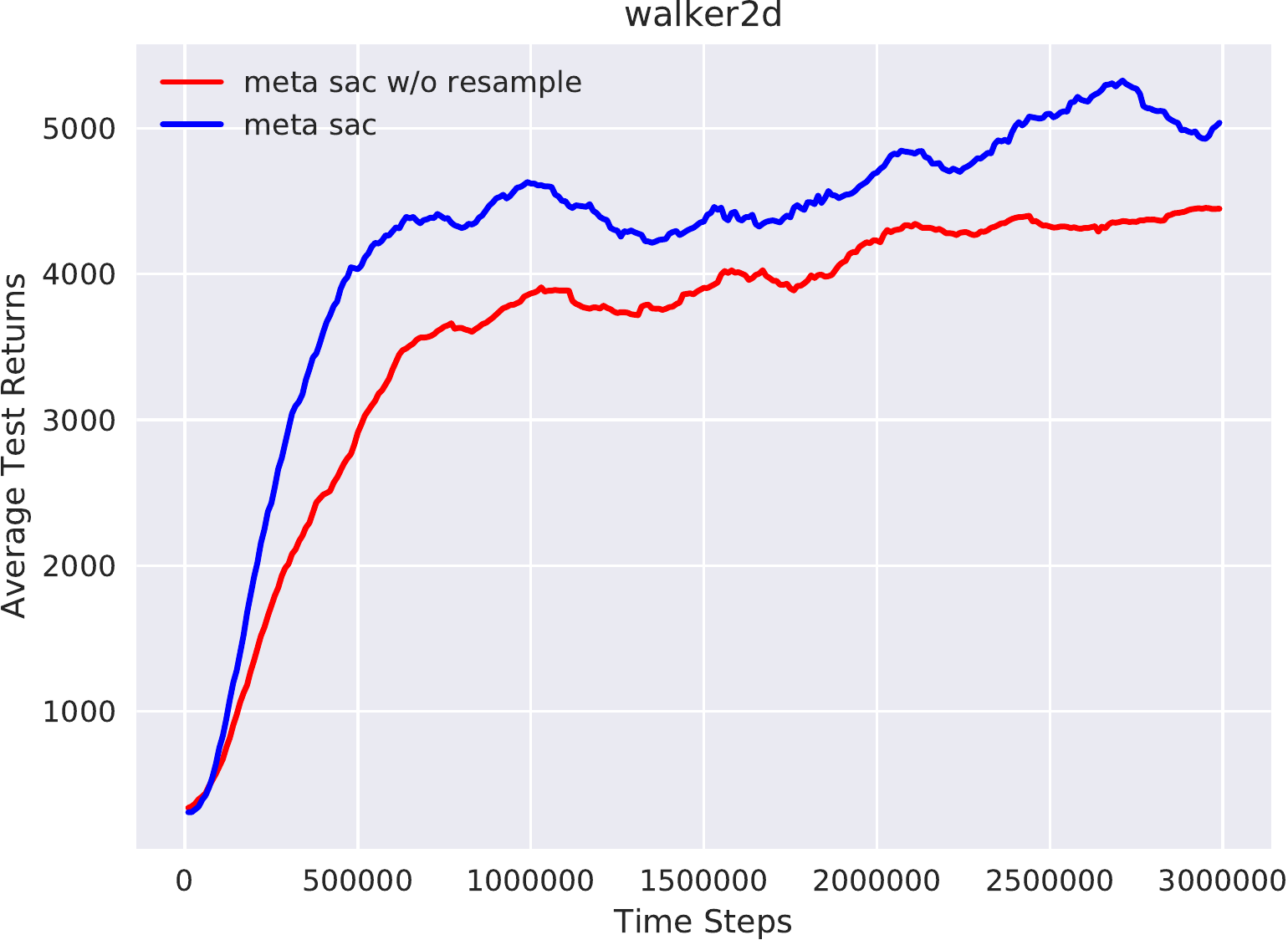} \\
    \multicolumn{3}{c}{(b) Ablation study on not using resampling (red curve) in the meta loss.}\\
    \includegraphics[width=0.32\textwidth]{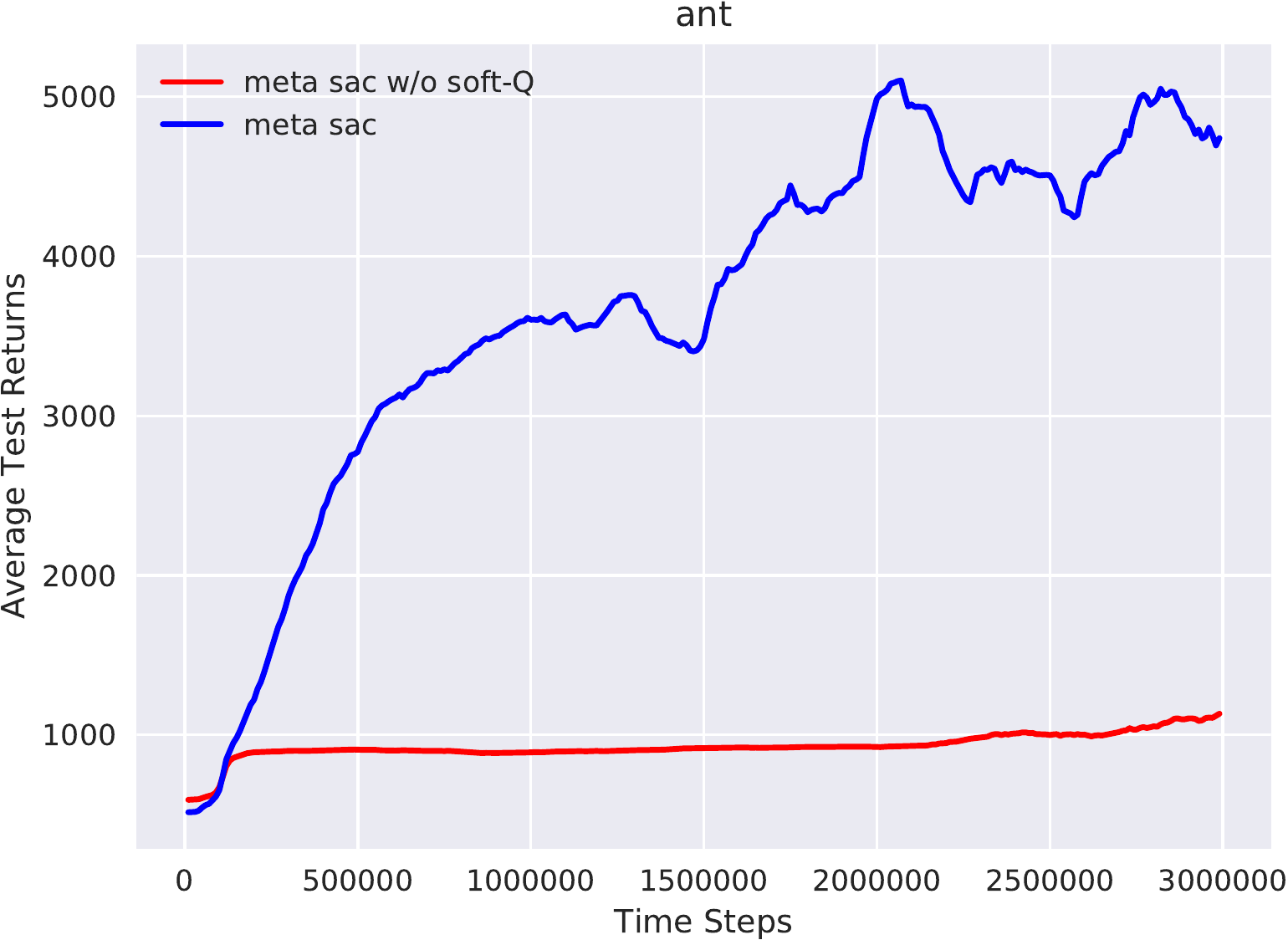} &
    \includegraphics[width=0.32\textwidth]{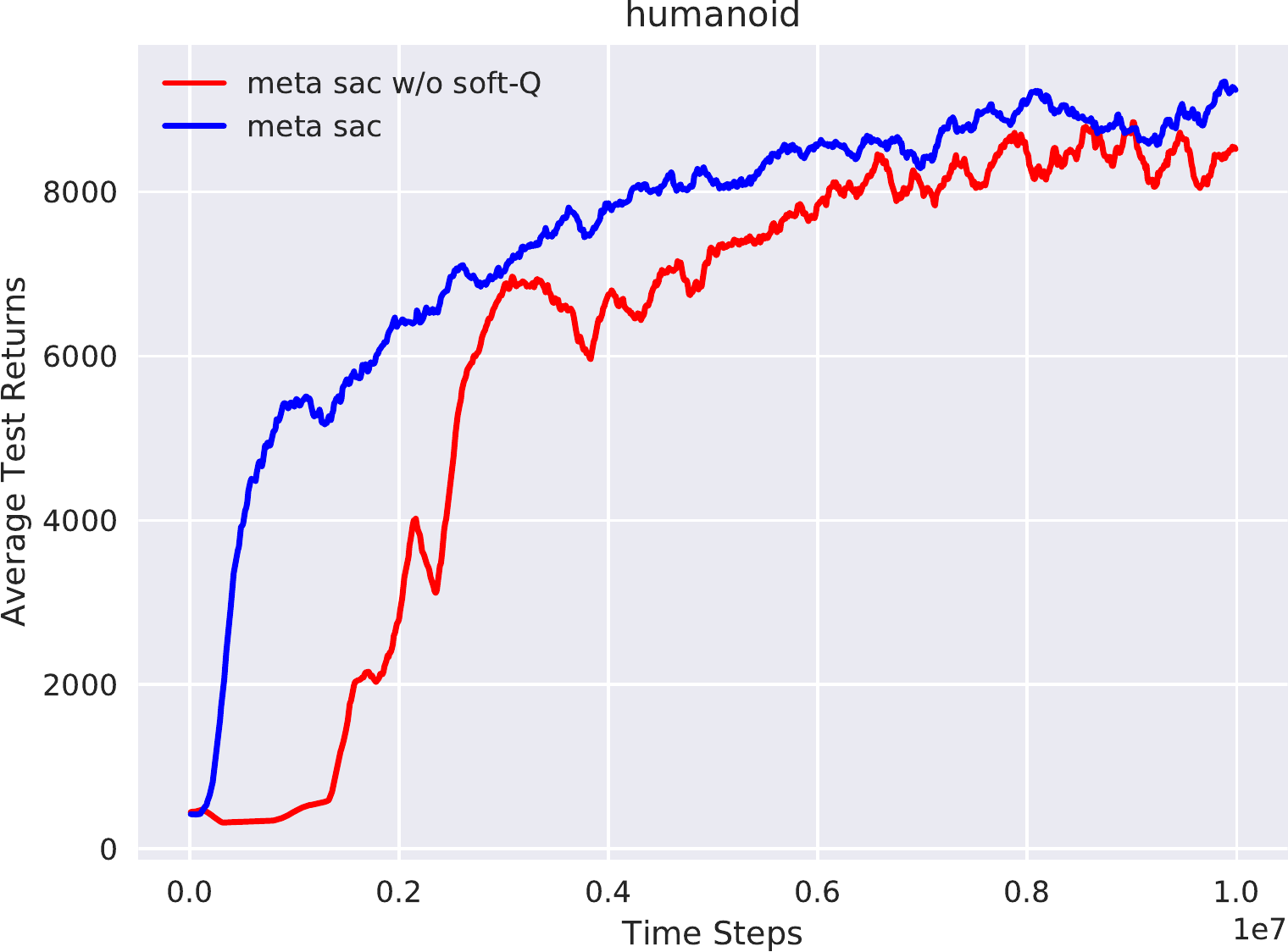} &
    \includegraphics[width=0.32\textwidth]{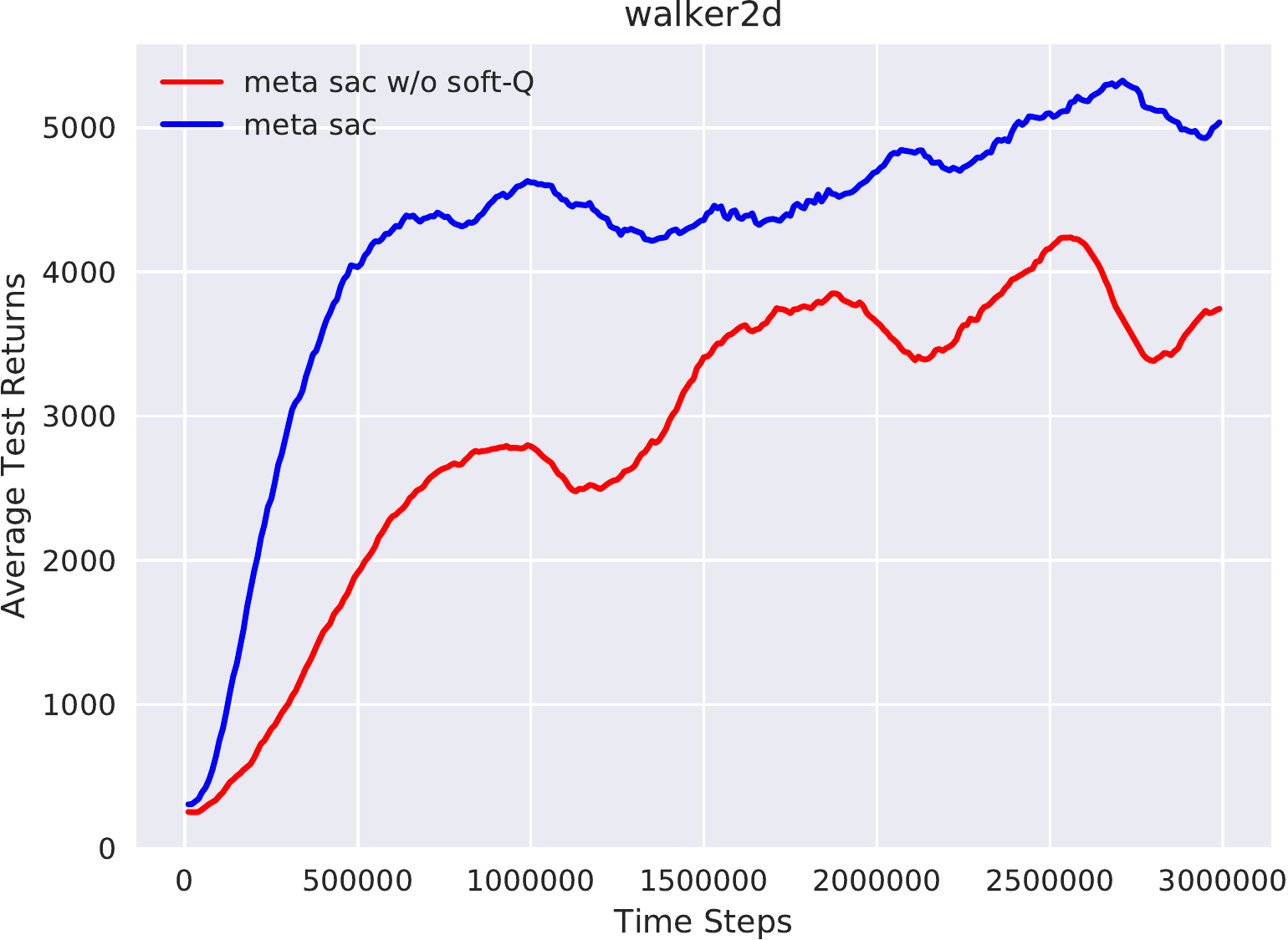} \\
    \multicolumn{3}{c}{(c) Ablation study on using classic Q (red curve) in the meta loss.}
    \end{tabular}
    \caption{Ablation studies on meta objective.}
    \label{fig:ablation}
\end{figure}

Finally, we test how the Meta-SAC performs if we change the soft-Q value used in Eq. \ref{eq:meta_alpha} to the classic Q-value. The result is shown in Figure \ref{fig:ablation} third row. As can be seen, when using the classic Q-function in the meta-loss, the performances degrades obviously in \texttt{Ant-v2} and \texttt{Walker2d-v2}. We suspect the reason is that such a meta loss discourages exploration and thus hurts the performance. 

\clearpage
\section{Ablation Study on Small Alpha in SAC-v1 on Humanoid}
\label{ab:sac_v1}
\begin{figure}[h]
    \centering
    \includegraphics[width=0.7\textwidth]{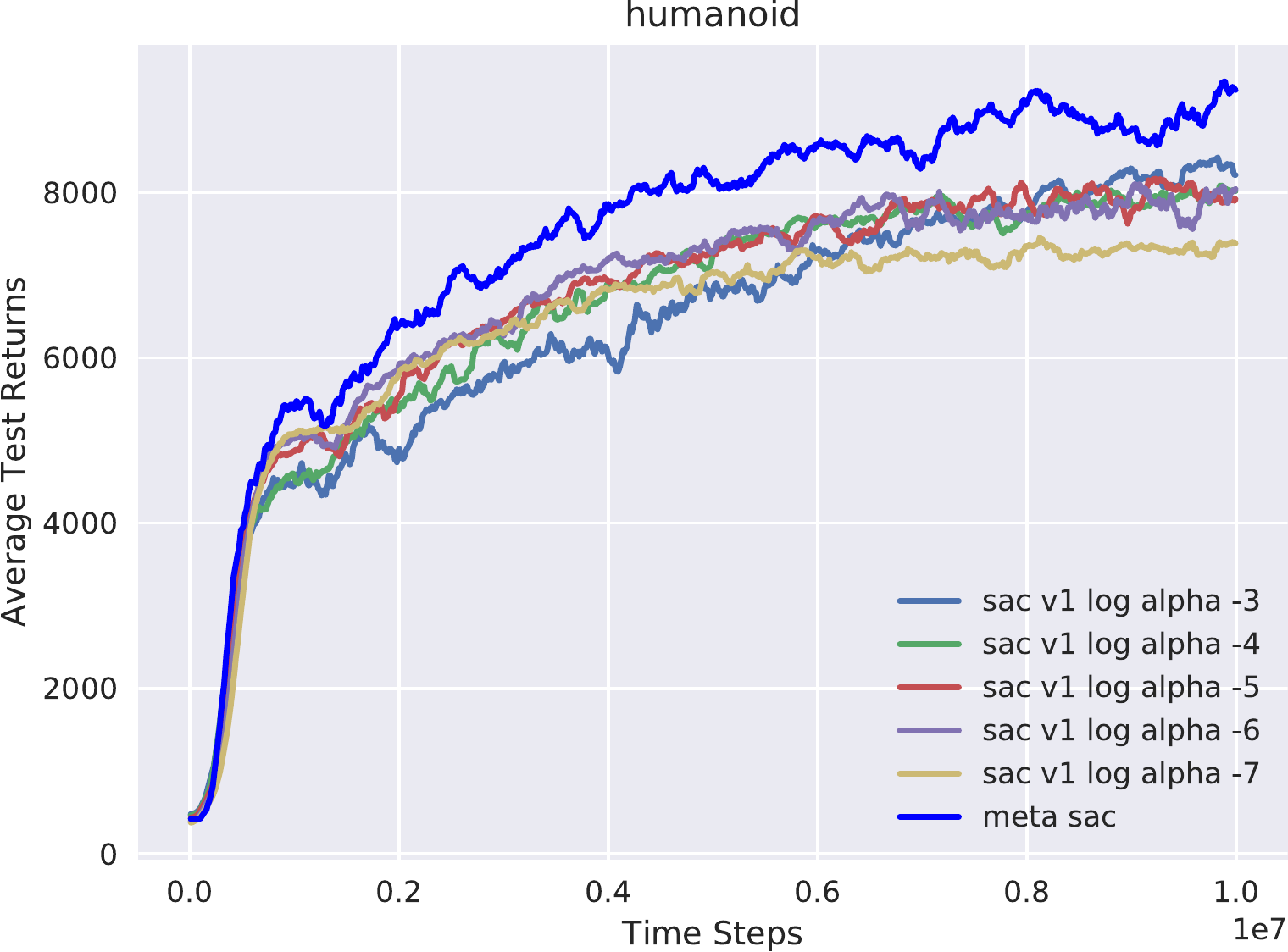}
    \caption{Ablation study on small alpha in SAC-v1 in Humanoid-v2.}
    \label{fig:sacv1}
\end{figure}

Figure \ref{fig:sacv1} shows the result of SAC-v1 with different \textit{fixed} alpha varying from $e^{-3}\approx 0.05$ to $e^{-7}\approx 0.001$ by grid search on log scale on \texttt{Humanoid-v2} task. We can see that simply decaying alpha will only worsen the final performance. Therefore, the early stage of meta-SAC where it maintains a high value of alpha to help exploration does contribute to its good performance.

\end{document}